\pdfoutput=1

\documentclass[11pt]{article}

\usepackage[final]{acl}

\usepackage{times}
\usepackage{latexsym}
\usepackage[T1]{fontenc}
\usepackage[utf8]{inputenc}
\usepackage{microtype}

\usepackage{inconsolata}

\usepackage{graphicx}

\usepackage[utf8]{inputenc} 
\usepackage[T1]{fontenc}    
\usepackage{hyperref}       
\usepackage{url}            
\usepackage{booktabs}       
\usepackage{amsfonts}       
\usepackage{nicefrac}       
\usepackage{microtype}      
\usepackage{xcolor}         

\usepackage{amsmath}
\usepackage{amssymb}
\usepackage{url}
\usepackage{booktabs} 
\usepackage{graphicx}
\usepackage{epstopdf}
\usepackage{subfigure}
\usepackage{verbatim}
\usepackage{bm}
\usepackage{array}
\usepackage{multirow}
\usepackage{subfigure}
\usepackage{makecell}
\usepackage{xcolor,colortbl}
\usepackage{xspace}
\usepackage{pifont}
\usepackage[edges]{forest}
\usepackage{float}
\usepackage{graphicx}
\usepackage{booktabs} 
\usepackage{amssymb} 
\usepackage{pifont} 
\usepackage{tabulary} 
\usepackage{tabularx} 

\definecolor{hidden-red}{RGB}{205, 44, 36}
\definecolor{hidden-blue}{RGB}{194,232,247}
\definecolor{hidden-orange}{RGB}{243,202,120}
\definecolor{hidden-green}{RGB}{34,139,34}
\definecolor{hidden-pink}{RGB}{255,245,247}
\definecolor{hidden-black}{RGB}{20,68,106}

\definecolor{Gray}{gray}{0.85}
\definecolor{LightCyan}{rgb}{0.88,1,1}

\newcolumntype{a}{>{\columncolor{Gray}}c}
\newcolumntype{b}{>{\columncolor{LightCyan}}c}

\usepackage{wrapfig}

\title{Causal Inference with Large Language Model: A Survey}

\author{Jing Ma \\
  Department of Computer and Data Sciences, Case Western Reserve University\\
  \texttt{jing.ma5@case.edu}
  }

\begin{document}
\maketitle
\begin{abstract}
Causal inference has been a pivotal challenge across diverse domains such as medicine and economics, demanding a complicated integration of human knowledge, mathematical reasoning, and data mining capabilities. Recent advancements in natural language processing (NLP), particularly with the advent of large language models (LLMs), have introduced promising opportunities for traditional causal inference tasks. This paper reviews recent progress in applying LLMs to causal inference, encompassing various tasks spanning different levels of causation. We summarize the main causal problems and approaches, and present a comparison of their evaluation results in different causal scenarios. Furthermore, we discuss key findings and outline directions for future research, underscoring the potential implications of integrating LLMs in advancing causal inference methodologies.
\end{abstract}

\section{Introduction}
\subsection{NLP, LLM, and Causality}
Causal inference is an important area to uncover and leverage the causal relationships behind observations, enabling a deep understanding of the underlying mechanism and potential interventions in real-world systems. Different from most classical statistical studies, causal inference presents unique challenges due to its focus on "causation instead of correlation", which intricates a complicated integration of human knowledge (e.g., domain expertise and common sense), mathematics, and data mining.   
Due to the inherent proximity to the human cognitive process, causal inference has become pivotal in many high-stakes domains such as healthcare \cite{glass2013causal}, finance \cite{atanasov2016shock}, and science \cite{imbens2015causal}. 

Traditional causal inference frameworks, such as structural causal model (SCM) \cite{pearl2009causality} and potential outcome framework \cite{imbens2015causal} have systematically defined causal concepts, quantities, and measures, followed up with multiple data-driven methods to discover the underlying causal relationships \cite{spirtes2016causal,nogueira2022methods,vowels2022d} and estimate the significance of causal effects \cite{winship1999estimation,yao2021survey}. Despite their success, existing causal methods still fall short of matching human judgment in several key areas, such as domain knowledge, logical inference, and cultural context \cite{kiciman2023causal, zevcevic2023causal, jin2023cladder}. 
Besides, most traditional causal inference approaches only focus on tabular data, lacking the ability to address causality in natural language. However, the demand for causal inference in natural language has persisted, offering numerous potential applications. For example, clinical text data from electronic health records (EHR) holds valuable causal information for healthcare research. Causal inference in NLP is a promising direction, offering both challenges and benefits. Advancements in large language models (LLMs) provide new opportunities to enhance traditional methods, bridging the gap between human cognition and causal inference \cite{feder2022causal}.



\begin{figure*}[t!]
\centering
\includegraphics[width=0.9\textwidth,height=2.4in]{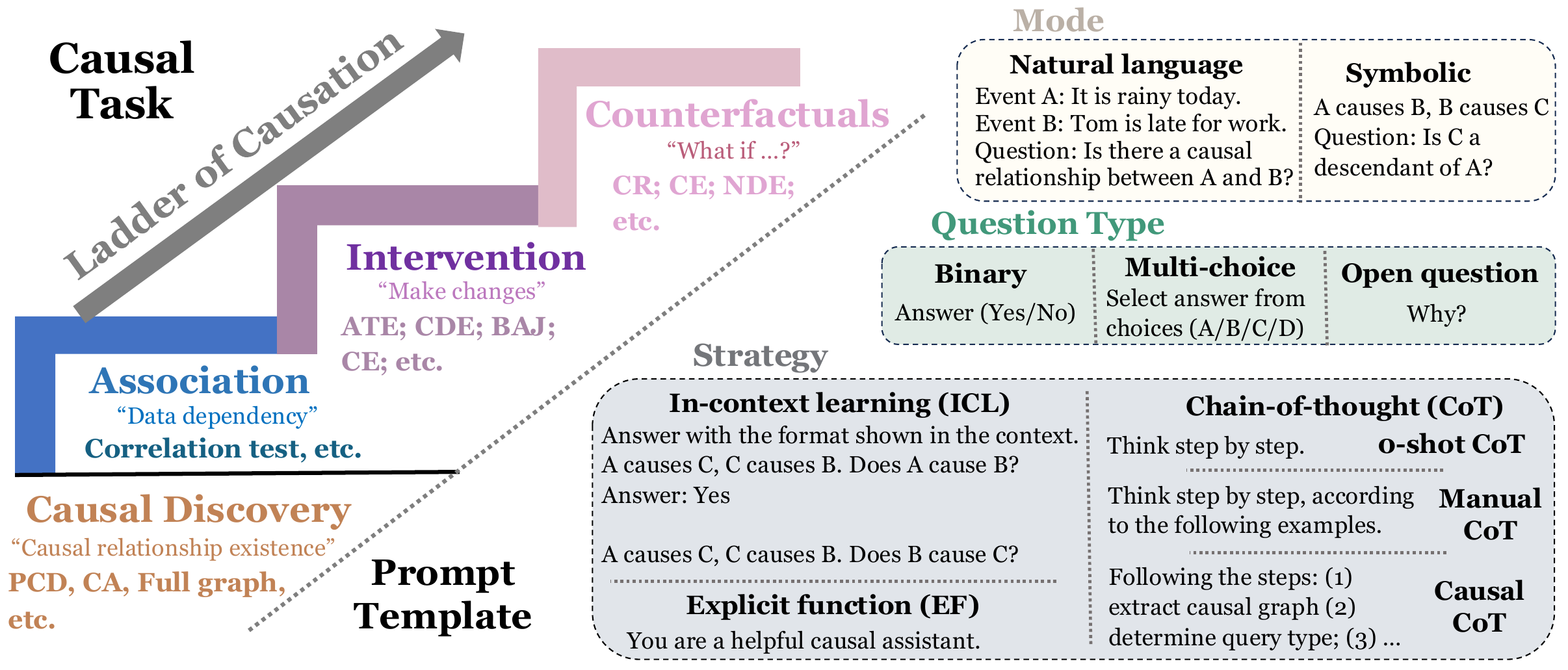}
\caption{Representative causal tasks, their positions in the causal ladder, and examples of prompts w.r.t. mode, question type, and prompting strategy. PCD = pairwise causal discovery; CA=causal attribution; ATE=average treatment effect; CDE=controlled direct effect; BAJ=backdoor adjustment; CE=causal explanation; CR=counterfactual reasoning; NDE=natural direct effect.}
\label{fig:causal_task}
\vspace{-2mm}
\end{figure*}

\subsection{Challenges of Causal Inference in NLP} 
Although LLMs have shown eye-catching success, causal inference poses unique challenges for LLM capabilities. Unlike regular data, natural language is unstructured, high-dimensional, and large-scale, making traditional causal methods ineffective. Besides, causal relationships in the text are often obscure, and the complex semantics require advanced language models to uncover them. These challenges create significant hurdles for causal tasks in NLP and require new approaches, presenting transformative opportunities for advancing causal inference research.

\vspace{-2mm}
\subsection{Benefits Brought by LLMs} 
Despite the challenges, the increasing sophistication of LLMs has enhanced their ability in causal inference from linguistic data. LLMs bring the following key benefits to causal inference:

\noindent \textbf{Domain knowledge.} Traditional causal methods focus on numerical data, but domain knowledge is crucial in fields like medicine for identifying causal relationships. LLMs can extract this knowledge from large-scale text, reducing dependence on human experts for causal inference.

\noindent \textbf{Common sense.} LLMs can capture human common sense, which aids causal reasoning across contexts. For instance, legal cases require logic, and common sense often identifies abnormal events as causes \cite{kiciman2023causal}.

\noindent \textbf{Sematic concept.} Natural language, with its nuances and complexity, presents challenges for identifying causal relationships. Advances in NLP and LLMs, particularly in semantic modeling, offer new opportunities for deeper causal analysis.

\noindent \textbf{Explainable causal inference.} LLMs can provide tools for more intuitive, natural language-based explanations of causal reasoning, making complex concepts more accessible and enhancing user interaction with causal inference results.

\subsection{Contribution and Uniqueness} 
\textbf{Contribution.} This survey systematically reviews existing research on using LLMs for causal inference with main contributions including: (1) We propose a clear categorization of studies, organized by tasks (Section \ref{sec:preliminary}) and technologies (Section \ref{sec:method}). 
(2) We present a detailed comparison of existing LLMs (Section \ref{sec:evaluation}) and highlight key insights, connections, and observations. 
(3) We provide a comprehensive summary of benchmark datasets, focusing on key aspects for further study (Table \ref{tab:benchmark_causal}). 
(4) We identify limitations and future research directions (Section \ref{sec:discussion}), offering new perspectives on underexplored areas and opportunities. 

\noindent\textbf{Differences from existing surveys.} Several previous surveys cover related topics \cite{liu2024large,kiciman2023causal}, while our survey differs as follows: (1) \textbf{Main scope:} We focus on "LLMs for causality," while other surveys with similar topic like \cite{liu2024large}, focus more on "causality for LLMs." (2) \textbf{Structure and content:} Our survey uniquely organizes research around tasks, methods, datasets, and evaluation, offering a clearer and more comprehensive review. (3) \textbf{Up-to-date:} We include the latest advancements, providing an up-to-date perspective on current trends and progress.

\tikzstyle{my-box}=[
    rectangle,
    draw=hidden-black,
    rounded corners,
    text opacity=1,
    minimum height=1.5em,
    minimum width=5em,
    inner sep=2pt,
    align=center,
    fill opacity=.5,
]
\tikzstyle{leaf}=[
    my-box, 
    minimum height=1.5em,
    text=black,
    align=left,
    font=\normalsize,
    inner xsep=4pt,
    inner ysep=4pt,
]
\tikzstyle{leaf_1}=[
    my-box, 
    minimum height=1.5em,
    fill=hidden-orange!60, 
    text=black,
    align=left,
    font=\normalsize,
    inner xsep=4pt,
    inner ysep=4pt,
]
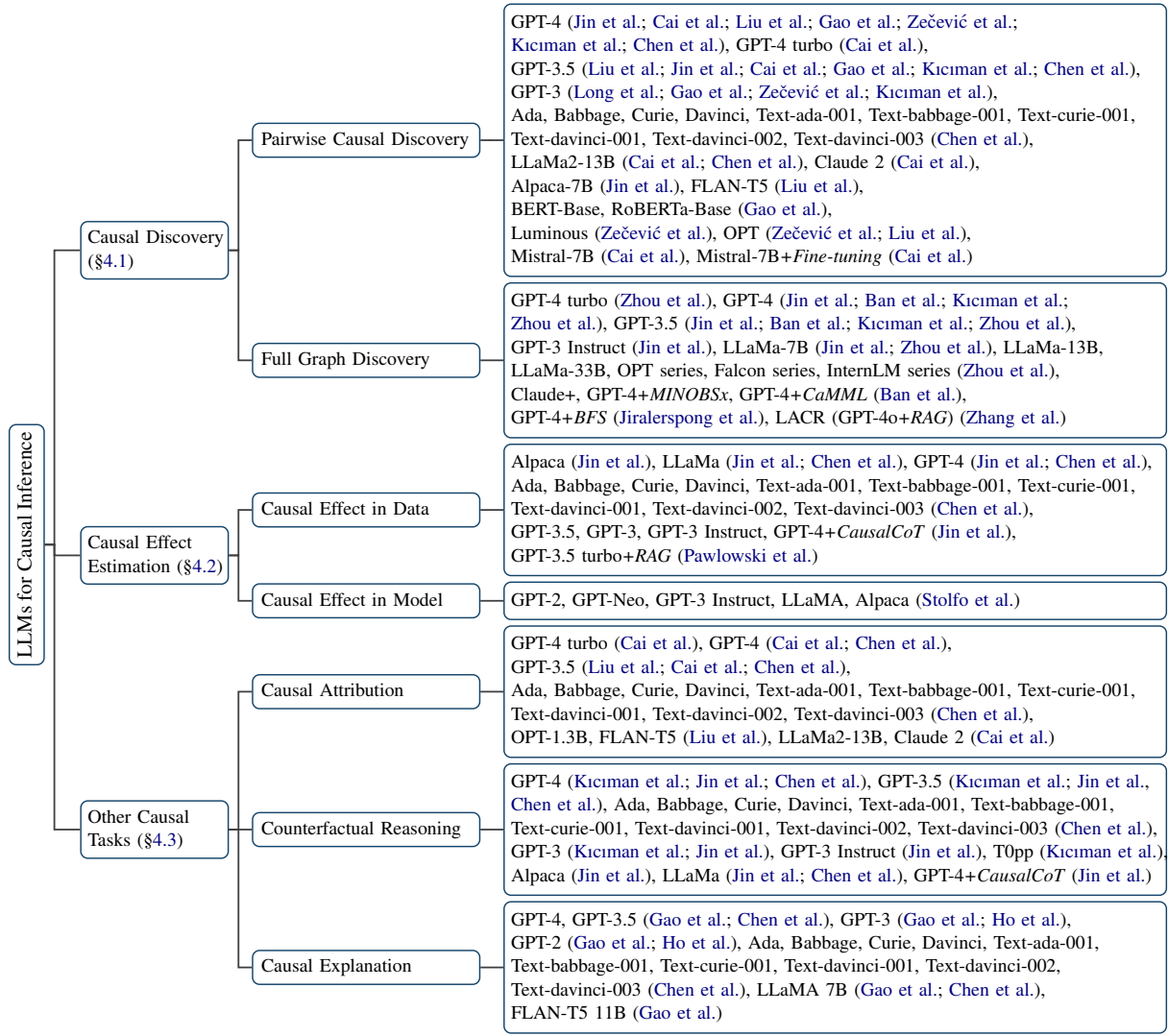
\begin{figure*}[t!]
    \centering
    \resizebox{\textwidth}{!}{
        \begin{forest}
            forked edges,
            for tree={
                grow=east,
                reversed=true,
                anchor=base west,
                parent anchor=east,
                child anchor=west,
                base=left,
                font=\large,
                rectangle,
                draw=hidden-black,
                rounded corners,
                align=left,
                minimum width=4em,
                edge+={darkgray, line width=1pt},
                s sep=3pt,
                inner xsep=4pt,
                inner ysep=3pt,
                line width=0.8pt,
                ver/.style={rotate=90, child anchor=north, parent anchor=south, anchor=center},
            },
            where level=1{
                text width=7.0em,
                font=\normalsize,
                inner xsep=4pt,
                }{},
            where level=2{
                text width=11.0em,
                font=\normalsize,
                inner xsep=5pt,
            }{},
            where level=3{
                text width=12em,
                font=\normalsize,
                inner xsep=4pt,
            }{},
            where level=4{
                text width=15em,
                font=\normalsize,
                inner xsep=4pt,
            }{},
            [
                LLMs for Causal Inference, ver
                [
                    Causal Discovery \\(\S \ref{sec:causal_discovery})
                    [
                        Pairwise Causal Discovery
                        [   
                            GPT-4 (\citeauthor{jin2023can,cai2023knowledge,liu2023trustworthy,gao2023chatgpt,zevcevic2023causal}{;}\\\citeauthor{kiciman2023causal,chen2024causal}){,}
                            GPT-4 turbo (\citeauthor{cai2023knowledge}){,} \\
                             GPT-3.5 (\citeauthor{liu2023trustworthy,jin2023can,cai2023knowledge,gao2023chatgpt,kiciman2023causal,chen2024causal}){,}\\
                             GPT-3 (\citeauthor{long2023can,gao2023chatgpt,zevcevic2023causal,kiciman2023causal}){,}  \\
                            Ada{,} 
                            Babbage{,}
                            Curie{,}
                            Davinci{,}
                            Text-ada-001{,} 
                            Text-babbage-001{,} Text-curie-001{,} \\
                            Text-davinci-001{,} 
                            Text-davinci-002{,}  Text-davinci-003 (\citeauthor{chen2024causal}){,}
                            \\LLaMa2-13B (\citeauthor{cai2023knowledge,chen2024causal}){,} Claude 2~(\citeauthor{cai2023knowledge}){,}  \\
                            Alpaca-7B (\citeauthor{jin2023can}){,} FLAN-T5 (\citeauthor{liu2023trustworthy}){,} Luminous (\citeauthor{zevcevic2023causal}){,} 
                            \\ BERT-Base{,} RoBERTa-Base (\citeauthor{gao2023chatgpt}){,} 
                            OPT (\citeauthor{zevcevic2023causal, liu2023trustworthy}){,}  \\
                            Mistral-7B (\citeauthor{cai2023knowledge}){,} Mistral-7B{\textit{\color{orange}{+Fine-tuning}}} (\citeauthor{cai2023knowledge}){,}\\
                            \textit{\color{orange}Causal agent} \cite{han2024causal}{,} \textit{\color{orange}Causal-Copilot} \cite{causalcopilot}
                                , leaf, text width=34em
                        ]
                    ]
                    [
                        Full Graph Discovery \\
                        [   
                            GPT-4 turbo (\citeauthor{zhou2024causalbench}){,} GPT-4 (\citeauthor{jin2023can,ban2023query,kiciman2023causal};\\
                            \citeauthor{zhou2024causalbench}){,} GPT-3.5 (\citeauthor{jin2023can,ban2023query,kiciman2023causal,zhou2024causalbench}){,} \\
                            GPT-3 Instruct (\citeauthor{jin2023can}){,} 
                            LLaMa-7B (\citeauthor{jin2023can,zhou2024causalbench}){,} LLaMa-13B{,} \\ LLaMa-33B{,} OPT series{,} Falcon series{,}  InternLM series (\citeauthor{zhou2024causalbench}){,} \\
                            Claude+{,} GPT-4\textit{\color{orange}{+MINOBSx}}{,} GPT-4\textit{\color{orange}{+CaMML}} (\citeauthor{ban2023query}){,} \\
                            GPT-4\textit{\color{orange}{+BFS}} (\citeauthor{jiralerspong2024efficient}){,} LACR (GPT-4o\textit{\color{orange}{+RAG}}) (\citeauthor{zhang2024causal}){,} \\
                            \textit{\color{orange}Causal agent} \cite{han2024causal}{,} \textit{\color{orange}Causal-Copilot} \cite{causalcopilot}
                                , leaf, text width=34em
                                , leaf, text width=34em
                        ]
                    ]
                ]
                [
                    Causal Effect \\ Estimation (\S \ref{sec:causal_effect})
                    [
                        Causal Effect in Data
                        [
                            Alpaca (\citeauthor{jin2023cladder}){,} LLaMa (\citeauthor{jin2023cladder,chen2024causal}){,}  GPT-4 (\citeauthor{jin2023cladder,chen2024causal}){,} \\
                            Ada{,} 
                            Babbage{,}
                            Curie{,}
                            Davinci{,}
                            Text-ada-001{,} 
                            Text-babbage-001{,} Text-curie-001{,} \\
                            Text-davinci-001{,} 
                            Text-davinci-002{,}  Text-davinci-003 (\citeauthor{chen2024causal}){,} \\
                              GPT-3.5{,} GPT-3{,} GPT-3 Instruct{,} 
                              GPT-4\textit{\color{orange}{+CausalCoT}} (\citeauthor{jin2023cladder}){,} \\GPT-3.5 turbo\textit{\color{orange}{+RAG}}{,} (\citeauthor{pawlowski2023answering}){,} \textit{\color{orange}Causal agent} \cite{han2024causal}
                                , leaf, text width=34em
                        ]
                    ]
                    [
                        Causal Effect in Model
                        [
                            GPT-2{,} GPT-Neo{,} GPT-3 Instruct{,} LLaMA{,} Alpaca (\citeauthor{stolfo2022causal})
                            , leaf, text width=34em
                        ]
                    ]
                ]
                [
                    Other Causal \\ Tasks (\S \ref{sec:other_causal})
                    [
                        Causal Attribution
                        [
                            GPT-4 turbo (\citeauthor{cai2023knowledge}){,} GPT-4  (\citeauthor{cai2023knowledge,chen2024causal}){,} \\
                            GPT-3.5 (\citeauthor{liu2023trustworthy,cai2023knowledge,chen2024causal}){,} \\ 
                            Ada{,} 
                            Babbage{,}
                            Curie{,}
                            Davinci{,}
                            Text-ada-001{,} 
                            Text-babbage-001{,} Text-curie-001{,} \\
                            Text-davinci-001{,} 
                            Text-davinci-002{,}  Text-davinci-003 (\citeauthor{chen2024causal}){,} \\
                            OPT-1.3B{,}
                            FLAN-T5 (\citeauthor{liu2023trustworthy}){,} 
                            LLaMa2-13B{,} Claude 2~(\citeauthor{cai2023knowledge})
                            , leaf, text width=34em
                        ]
                    ]
                    [
                        Counterfactual Reasoning
                        [
                            GPT-4 (\citeauthor{kiciman2023causal,jin2023cladder,chen2024causal}){,} GPT-3.5 (\citeauthor{kiciman2023causal,jin2023cladder}{,}\\
                            \citeauthor{chen2024causal}){,} 
                            Ada{,} 
                            Babbage{,}
                            Curie{,}
                            Davinci{,}
                            Text-ada-001{,} 
                            Text-babbage-001{,} \\ Text-curie-001{,} 
                            Text-davinci-001{,} 
                            Text-davinci-002{,}  Text-davinci-003 (\citeauthor{chen2024causal}){,} \\
                            GPT-3 (\citeauthor{kiciman2023causal,jin2023cladder}){,} GPT-3 Instruct (\citeauthor{jin2023cladder}){,} T0pp (\citeauthor{kiciman2023causal}){,} \\
                            Alpaca (\citeauthor{jin2023cladder}){,} LLaMa (\citeauthor{jin2023cladder,chen2024causal}){,} GPT-4\textit{\color{orange}{+CausalCoT}} (\citeauthor{jin2023cladder})
                            , leaf, text width=34em
                        ]
                    ]
                    [
                        Causal Explanation
                        [
                            GPT-4{,} GPT-3.5 (\citeauthor{gao2023chatgpt,chen2024causal}){,} GPT-3 (\citeauthor{gao2023chatgpt,ho2022wikiwhy}){,} \\
                            GPT-2 (\citeauthor{gao2023chatgpt,ho2022wikiwhy}){,}   
                            Ada{,} 
                            Babbage{,}
                            Curie{,}
                            Davinci{,} 
                            Text-ada-001{,} \\
                            Text-babbage-001{,} Text-curie-001{,} 
                            Text-davinci-001{,} 
                            Text-davinci-002{,}  \\ 
                            Text-davinci-003 (\citeauthor{chen2024causal}){,} 
                            LLaMA 7B (\citeauthor{gao2023chatgpt,chen2024causal}){,} \\FLAN-T5 11B (\citeauthor{gao2023chatgpt}){,} \textit{\color{orange}Causal-Copilot} \cite{causalcopilot}
                            , leaf, text width=34em]
                    ]
                ]
            ]
        \end{forest}
    }
    \caption{The major causal tasks and LLMs evaluated for these tasks. \thanks{Noticeably, the citations in the figure correspond to the work of evaluations, rather than the original work of these models themselves. The strategies that are \textit{not} merely based on prompting are highlighted in \color{orange}{orange.}}}
    \label{fig:taxonomy_causal}
    \vspace{-2mm}
\end{figure*}

\section{Preliminaries}
\label{sec:preliminary}
\subsection{Causality}

\noindent\textbf{Structural causal model.} Structural causal model (SCM) \cite{pearl2009causality} is a widely used model to describe the causal relationships inside a system. An SCM is defined with a triple $(U, V, F)$: $U$ is a set of exogenous variables, whose causes are out of the system; $V$ is a set of endogenous variables, which are determined by variables in $U\cup V$;
 $F=\{f_1(\cdot),f_2(\cdot),...,f_{|V|}(\cdot)\}$ is a set of functions (a.k.a. \textit{structural equations}). For each $V_i\in V$, $ V_i = f_i(pa_i,U_{i})$, where ``$pa_i\subseteq V \setminus{V_i}$” and ``$U_{i}\subseteq U$" are variables that directly cause $V_i$. 
Each SCM is associated with a \textit{causal graph}, which is a directed acyclic graph (DAG), where each node stands for
a variable, and each arrow is a causal relationship. 

\noindent\textbf{Ladder of causation.} The ladder of causation \cite{pearl2018book,bareinboim2022pearl} defines three rungs (Rung 1: \textit{Association}; Rung 2: \textit{Intervention}; Rung 3: \textit{Counterfactuals}) to describe different levels of causation. Each higher rung indicates a more advanced level of causality. The first rung "Association" involves statistical dependencies, related to questions like "What is the correlation between taking a medicine and a disease?". Rung 2 "Intervention" moves further to allow interventions on variables, with questions like "If I take a certain medicine, will my disease be cured?". Rung 3 "Counterfactuals" relates to imagination or retrospection queries like "What if I had acted differently?", "Why?". Answering such questions requires knowledge related to the corresponding SCM. Counterfactual ranks the highest because it subsumes the first two rungs. A model that can handle counterfactual queries can also handle associational and interventional queries. 

\subsection{Causal Tasks and Related Rungs in Ladder of Causation}
Causal inference involves various tasks. Figure \ref{fig:causal_task} shows an overview of these tasks and their positions in the ladder of causation. We also show some examples of prompts w.r.t. mode, question type, and prompting strategy. We list several main causal tasks as follows:

\noindent \textbf{Causal discovery}. Causal discovery aims to infer causal relationships from data. It includes discovering the causal graph and the structural equations associated with these causal relationships. Although causal discovery is not explicitly covered in the ladder of causation, it is often considered as "Rung 0" as it serves as a fundamental component in causal inference. Typical causal discovery questions include \textit{pairwise causal discovery (PCD)} that only focuses on a pair of variables, and \textit{full graph discovery} involving variables in the whole data system.

\noindent \textbf{Causal effect estimation}. Causal effect estimation (a.k.a. treatment effect estimation) aims to quantify the strength of the causal influence of a particular intervention or treatment on an outcome. 
In different scenarios, researchers may focus on the causal effect of different granularities, such as \textit{individual treatment effect} (\textit{ITE}, i.e., treatment effect on a specific individual), \textit{conditional average treatment effect} (\textit{CATE}, i.e., average treatment effect on a certain subgroup of population), \textit{average treatment effect on the treated group} (\textit{ATT}), and \textit{average treatment effect} (\textit{ATE}, i.e., average treatment effect on the entire population). Besides, people are also interested in the direct/indirect causal effects in certain scenarios, such as \textit{natural direct effect} (\textit{NDE}), \textit{controlled direct effect} (\textit{CDE}), and \textit{natural indirect effect} (\textit{NIE}). Another task related to causal effect estimation is \textit{backdoor adjustment (BAJ)}, which aims to block all backdoor paths \cite{pearl2009causality} from the treatment to the outcome to exclude non-causal associations. Causal effect estimation tasks often span over Rung 2 and Rung 3.  

\noindent \textbf{Other tasks}. There are many other tasks in causal inference. Among them, \textbf{causal attribution (CA)} refers to the process of attributing a certain outcome to certain events. \textbf{Counterfactual reasoning (CR)} investigates what might have happened if certain events or conditions had been different from what actually occurred. It explores hypothetical scenarios by considering alternative outcomes based on changes in ``what if" circumstances. \textbf{Causal explanation (CE)} aims to generate explanations for an event, a prediction, or any causal reasoning process. This task often needs to answer causal questions in a specified human-understandable form or plain language. It is often in Rung 2 or 3, depending on the specific context. 
It is worth noting that, in many cases, different causal tasks may exhibit natural overlap in their scope. For instance, causal attribution and explanation commonly intersect with causal discovery and causal effect estimation. However, each task maintains a distinct focus.

\begin{table*}[t!]
\centering
\small
\begin{tabular}{llccccc}
\toprule
\textbf{Dataset} & \textbf{Task} & \textbf{Size (Unit)} & \textbf{Domain} & \textbf{Real} & \textbf{\# of sources} & \textbf{Citations} \\
\midrule
\textbf{CEPairs} \citeyearpar{mooij2016distinguishing}  & CD &  108 (P) & Meteorology, etc. & R & 37 &\citeyearpar{mooij2016distinguishing,cai2023knowledge,zevcevic2023causal,kiciman2023causal} \\
\textbf{Sachs} \citeyearpar{zheng2024causal}  & CD& 20 (R)& Biology & R & 1& \citeyearpar{cai2023knowledge,zheng2024causal,zhang2024causal} \\
\textbf{Corr2Cause} \citeyearpar{jin2023can}  & CD & 200K (S)& Math & S & 1& \citeyearpar{jin2023can}\\
\textbf{CLADDER} \citeyearpar{jin2023cladder}  & Eff, CR, CE & 10K (S) & Dailylife, etc.& S & 1&\citeyearpar{jin2023cladder,jin2023can} \\
\textbf{BN Repo} \citeyearpar{BNRepo}  & CD& 4$\sim$84 (R)& Health, etc. & R& 8&\citeyearpar{ban2023query}\\
\textbf{COPA} \citeyearpar{roemmele2011choice}  & CD & 1,000 (Q) & Commonsense& R& 1&\citeyearpar{gao2023chatgpt,roemmele2011choice} \\
\textbf{E-CARE} \citeyearpar{du2022care}  & CD, CE & 21K (Q) & Commonsense & R & 1&\citeyearpar{gao2023chatgpt,du2022care} \\
\textbf{Asia} \citeyearpar{lauritzen1988local}  & CD & 8 (R) & Health & R & 1 & \citeyearpar{lauritzen1988local,jiralerspong2024efficient,zhang2024causal}\\
\textbf{CausalNet} \citeyearpar{luo2016commonsense}  & CD & 62M (R) & Web text & S & 1&\citeyearpar{luo2016commonsense,du2022care}\\
\textbf{CausalBank} \citeyearpar{li2021guided}		& CD	& 314 M	(P)	& Web text	& S & 1&\citeyearpar{li2021guided, du2022care}	\\		
\textbf{WIKIWHY} \citeyearpar{ho2022wikiwhy}  & CD,CE & 9K (Q) & Wikipedia & R & 1&\citeyearpar{ho2022wikiwhy}\\
\textbf{Neuro Pain} \citeyearpar{tu2019neuropathic}  & CD & 770 (R) & Health & S & 1&\citeyearpar{tu2019neuropathic,kiciman2023causal,tu2023causal}\\
\textbf{Arctic Ice} \citeyearpar{huang2021benchmarking}  & CD & 48 (R) & Climate & R & 1&\citeyearpar{huang2021benchmarking,kiciman2023causal} \\
\textbf{CRASS} \citeyearpar{frohberg2022crass}  & CR & 275 (Q) & Commonsense & R & 1&\citeyearpar{frohberg2022crass}\\
\textbf{CausalQA} \citeyearpar{bondarenko2022causalqa}  & CD, CE & 1.1M (Q)& Web text, etc. & R & 10&\citeyearpar{blubaum2024causal,bondarenko2022causalqa,tao2024comprehensive}\\
\textbf{CALM-Bench} \citeyearpar{dalal2023calm}  & CD, CA & 281K (Q) & Science, etc. & R & 6 &\citeyearpar{dalal2023calm} \\
\textbf{CausalBench} \citeyearpar{zhou2024causalbench}  & Corr, CD & 4$\sim$195 (R)& Health, etc.  &R & 15 & \citeyearpar{zhou2024causalbench}\\
\textbf{CaLM} \citeyearpar{chen2024causal}  & Rung 1$\sim$3 & 126K (S) & Commonsense, etc. & R \& S & 20& \citeyearpar{chen2024causal} 
\\
\bottomrule
\end{tabular}
\caption{Datasets for LLM-related causal inference, with publication year, applicable tasks (CD=causal discovery; Eff=effect estimation; CR=counterfactual reasoning; CE=causal explanation), dataset size (as these datasets are not in a consistent form, we show the size w.r.t. different units, where P=causal pairs; R=causal relations; S=samples; Q=questions), domain, generation process (R: real-world; S: synthetic), number of data sources, and citations.}
\label{tab:benchmark_causal}
\vspace{-4mm}
\end{table*}

\section{Methodologies}
\label{sec:method}
\vspace{-2mm}
Recent efforts \citep{kiciman2023causal,chen2024causal,gao2023chatgpt} have explored leveraging LLMs for causal tasks. Unlike traditional data-driven or expert knowledge-based approaches, LLMs introduce novel methodologies, offering new perspectives for discovering and utilizing causal knowledge. Figure \ref{fig:taxonomy_causal} lists LLMs developed or evaluated for causal tasks. We categorize current LLM methodologies for causal tasks as follows:

\noindent\textbf{Prompting.} Most existing works on causal reasoning with LLMs \cite{chen2024causal,kiciman2023causal,long2023can,jin2023cladder} focus on prompting, as it is the simplest approach. This includes regular strategies (like basic prompting, In-Context Learning (ICL) \cite{brown2020language}, and Chain-of-Thought (CoT) \cite{wei2022chain}) and causality-specific strategies \cite{jin2023cladder}. For regular prompting, basic prompts (i.e., directly describing the question without any example or instruction) are most frequently used. There are also other efforts to devise more advanced prompting strategies.
Among them, CaLM \cite{chen2024causal} has tested 9 prompting strategies including basic prompt, adversarial prompt \cite{wallace2019universal,perez2022ignore}, ICL, 0-shot CoT (e.g., \textit{"let’s think step by step"} without any examples) \cite{kojima2022large}, manual CoT (i.e., guide models with manually designed examples), and explicit function (EF) (i.e., using encouraging language in prompts) \cite{chen2024causal}.  Other works \cite{kiciman2023causal,long2023can,gao2023chatgpt,ban2023query} also design different prompt templates. These works show substantial improvement potential of prompt engineering in causal reasoning tasks. For example, studies \cite{kiciman2023causal,chen2024causal,long2023can} highlight that simple phrases like \textit{"you are a helpful causal assistant"} can significantly boost performance. Additionally, there are causality-specific strategies, such as CausalCoT \cite{jin2023cladder}, which combine CoT prompting with causal inference principles \cite{pearl2018book}. Prompting-based methods can quickly and flexibly adapt to different causal tasks, but are still limited by the specificity of the prompt and thus easily lead to inconsistent causal responses.

\noindent\textbf{Fine-tuning.} Fine-tuning, as a widely recognized technique in general LLMs, is now also starting to gain attention for its application in causal tasks. \citet{cai2023knowledge} propose a fine-tuned LLM for the pairwise causal discovery task (PCD, introduced in Section \ref{sec:causal_discovery}). This method generates a fine-tuning dataset with a Linear, Non-Gaussian, Acyclic Model \cite{shimizu2006linear}, uses Mistral-7B-v0.2 \cite{jiang2023mistral} as LLM backbone, and runs instruction finetuning with LoRA \cite{hu2021lora}. The results achieve significant improvement compared with the backbone without fine-tuning. However, effective fine-tuning requires large computational resources, and may also suffer from overfitting problems.

\noindent\textbf{Combining traditional causal methods.} A line of studies combines LLMs with traditional causal methods. Considering causal inference often heavily relies on numerical reasoning, an exploration in \citet{ban2023query} leverages LLMs and data-driven causal algorithms such as MINOBSx \cite{li2018bayesian} and CaMML \cite{o2006causal}. This method outperforms both original LLMs and data-driven methods, indicating a promising future for combining the language understanding capability of LLMs and the numerical reasoning skills of data-driven methods in complicated causal tasks.  Jiralerspong et al. \cite{jiralerspong2024efficient} combine LLM with a breadth-first search (BFS) approach for full graph discovery. It considers each causal relation query as a node expansion process, and gradually constructs the causal graph by traversing it with BFS. This method significantly reduces the time complexity from $O(n^2)$ to $O(n)$, where $n$ is the number of variables. While it does not require access to observational data, their experiments show that the performance can be further enhanced with observational statistics. Recently, there have been increasing efforts on causality-driven LLM-based agents such as Causal Agent \cite{han2024causal} and Causal-Copilot \cite{causalcopilot} that automatically use LLMs to invoke causal tools for causal problems. 
This line of methods can potentially answer more complex causal questions than traditional methods and plain LLMs, but harmonizing between LLMs and data-driven causal approaches can also be a subtle problem.

\begin{table*}[t]
\centering
\begin{tabular}{l c cc cc a b}
\toprule
\multirow{2}{*}{\textbf{Model}} & \textbf{CEPairs} & \multicolumn{2}{c}{\textbf{E-CARE}} & \multicolumn{2}{c}{\textbf{COPA}} & \textbf{CALM-CA} & \textbf{Neuro Pain}\\
\cmidrule(lr){2-2} \cmidrule(l){3-4} \cmidrule(l){5-6} \cmidrule(l){7-7} \cmidrule(l){8-8}
 & Binary & Choice & Binary & Choice & Binary &  Binary & Choice  \\
\midrule
ada          &  0.50    &  0.48 &   0.49 &  0.49 &  0.49 & 0.57 & 0.40\\
text-ada-001 &  0.49    &  0.49 &   0.33 &  0.50 &  0.35 & 0.48 & 0.50\\
Llama2 (7B)  &  -       &  0.53 &   0.50 &  0.41 &  0.35 & 0.32 & -\\
Llama2 (13B) &  -       &  0.52 &   0.50 &  0.44 &  0.36 & 0.42 & -\\
Llama2 (70B) &  -       &  0.52 &   0.44 &  0.50 &  0.45 & 0.49 & -\\
Qwen (14B) & - & 0.66 & 0.52 & 0.77 & 0.53 & 0.52 & - \\
babbage     &   0.51    &  0.49 &   0.36 &  0.49 &  0.40 & 0.58& 0.50\\
text-babbage-001& 0.50  &  0.50 &   0.50 &  0.49 &  0.50 & 0.56& 0.51\\
curie           & 0.51  &  0.50 &   0.50 &  0.50 &  0.50 & 0.58& 0.50\\
text-curie-001  & 0.50  &  0.50 &   0.50 &  0.51 &  0.50 & 0.58& 0.50\\
davinci         & 0.48  &  0.50 &   0.49 &  0.50 &  0.51 & 0.58& 0.38\\
text-davinci-001& 0.50  &  0.50 &   0.50 &  0.50 &  0.50 & 0.52& 0.50\\
text-davinci-002& 0.79  &  0.66 &   0.64 &  0.80 &  0.67 & 0.69& 0.52\\
text-davinci-003& 0.82  &  0.77 &   0.66 &  0.90 &  0.77 & 0.80& 0.55\\
GPT-3.5-Turbo   & 0.81  &  \textbf{0.80} &   0.66 & \textbf{0.92} & 0.66 & 0.72 & 0.71\\
GPT-4           & -  &  0.74 &   \textbf{0.68} & 0.90 & \textbf{0.80} & \textbf{0.93} & \textbf{0.78}\\
\midrule
\midrule
GPT-4 (0-shot ICL)   & - & 0.83 & 0.71 & 0.97 & 0.78 & 0.90 & -\\
GPT-4 (1-shot ICL)   & - & 0.81 & 0.70 & 0.93 & 0.76 & 0.90 & -\\
GPT-4 (3-shot ICL)   & - & 0.71 & 0.70 & 0.80 & 0.81 & 0.91 & -\\
GPT-4 (0-shot CoT)          & - & 0.77 & 0.68 & 0.91 & 0.79 & 0.92 & -\\ 
GPT-4 (Manual CoT)          & - & 0.79 & \textbf{0.73} & 0.97 & \textbf{0.82} & \textbf{0.95} & -\\  
GPT-4 (EF)           & - & \textbf{0.83} & 0.71 & \textbf{0.98} & 0.80 & 0.92 & \textbf{0.84}\\
\bottomrule
\end{tabular} 
\caption{Performance (accuracy) of LLMs for causal discovery. Datasets include CausalEffectPairs (CEpairs), E-CARE, COPA, CALM-CA, and Neuro Pain. The white columns evaluate LLMs on PCD; in the \colorbox{lightgray}{gray} column, on causal attribution; and in the \colorbox{LightCyan}{cyan} column, on full graph discovery. The upper part shows results with basic prompts; the lower part shows GPT-4 results with different prompting strategies. The tasks are shown in either binary "yes/no" or multi-choice formats. Results are drawn from \citet{kiciman2023causal} and \citet{chen2024causal}.}
\vspace{-3mm}
\label{table:causal_discovery_performance}
\end{table*}

\noindent\textbf{Knowledge augmentation.} LLMs with augmented knowledge can often better execute tasks for which they are not well-suited
to perform by themselves, particularly for causal tasks that require domain-specific knowledge. Pawlowski et al. \citeyearpar{pawlowski2023answering} introduced two types of knowledge augmentation: context augmentation, which provides causal graphs and ITEs in the prompt, and tool augmentation, offering API access to expert systems for causal reasoning. Tool augmentation performs more robustly across varying problem sizes, as the LLM relies on the API for reasoning instead of reasoning through the graph itself. LACR \cite{zhang2024causal} applies retrieval augmented generation (RAG) to enhance the knowledge base of LLM for causal discovery, where the knowledge resources are from a large scientific corpus containing hidden insights about associational/causal relationships. Similar approaches \cite{samarajeewa2024causal} employ causal graphs as external sources for causal reasoning. Knowledge augmentation is especially effective for domain-specific tasks, as seen in RC$^2$R \cite{yu2024fusing} for financial risk analysis and CausalKGPT \cite{zhou2024causalkgpt} for aerospace defect analysis. However, high-quality causal knowledge sources are hard to collect, which can limit the LLM performance and also increase the method complexity.

\vspace{-1mm}
\section{Evaluations of LLMs in Causal Tasks}
\vspace{-1mm}
\label{sec:evaluation}

This section summarizes recent evaluation results of LLMs in causal tasks. We mainly focus on causal discovery and causal effect estimation, and also introduce several representative tasks spanning Rung 1$\sim$3. A collection of datasets used in LLM-related causal tasks is shown in Table \ref{tab:benchmark_causal}. In Table \ref{table:causal_discovery_performance} and Table \ref{table:causal_effect_performance}, we compare the performance of different LLMs in different tasks (including causal discovery and other tasks spanning different rungs) on multiple datasets. The mentioned LLMs include ada, babbage, curie, davinci \cite{brown2020language}, Qwen (14B) \cite{bai2023qwen}, text-ada-001, text-babbage-001, text-curie-001, text-davinci-001, text-davinci-002, text-davinci-003 \cite{ouyang2022training},
Llama 2 (7B, 13B, 70B) \cite{touvron2023llama}, OpenAI's GPT series \cite{achiam2023gpt,openai2022chatgpt}, Mistral (7B) \cite{jiang2023mistral}, and Claude 2 \cite{Claude2}.

\subsection{LLM for Causal Discovery}
%
\label{sec:causal_discovery}

\begin{table}[]
\centering
\begin{tabular}{lc}
\toprule
\textbf{Model}                     & \textbf{Accuracy}  \\
\midrule
GPT-3.5                   & 0.58 \\
GPT-4                     & 0.76 \\
GPT-4-Turbo               & 0.82 \\
Llama2 (13B)                & 0.74 \\
Claude 2                  & 0.74 \\
Mistral (7B) v0.2           & 0.75 \\
Fine-tuned Mistral (7B) v0.2 & \textbf{0.90}\\
\bottomrule
\end{tabular}
\caption{Accuracy of LLMs and a fine-tuned Mistral in PCD. The results are sourced from \citet{cai2023knowledge}.}
\label{table: finetuning}
\end{table}

\begin{table*}[t]
\centering
\begin{tabular}{l a cc cb b bb}
\toprule
\multirow{2}{*}{\textbf{Model}} & \textbf{CLADDER} & \multicolumn{3}{c}{\textbf{CaLM}} & \textbf{CLADDER} & \textbf{CaLM} & \textbf{CRASS} & \textbf{E-CARE}\\
\cmidrule(lr){2-2} \cmidrule(l){3-5} \cmidrule(l){6-6}  \cmidrule(l){7-7} \cmidrule(l){8-8} \cmidrule(l){9-9}
& Corr & ATE & CDE & BAJ & CR & NDE & CR & CE\\
\midrule
ada             & 0.26 & 0.02 & 0.03 & 0.13 & 0.30 & 0.05 & 0.26 & 0.22\\
text-ada-001    & 0.25 & 0.01 & 0.01 & 0.29 & 0.28 & 0.01 & 0.24 & 0.33\\
Llama2 (7B)     & 0.50 & 0.03 & 0.02 & 0.18 & 0.51 & 0.03 & 0.11 & 0.42\\
Llama2 (13B)    & 0.50 & 0.01 & 0.01 & 0.19 & 0.52 & 0.02 & 0.20 & 0.39\\
Llama2 (70B)    & 0.51 & 0.09 & 0.09 & 0.13 & 0.52 & 0.13 & 0.17 & 0.42\\
Qwen (14B) & 0.45 & 0.12 & 0.12 & 0.30 & 0.39 & 0.10 & 0.34 & 0.39\\
babbage         & 0.39 & 0.03 & 0.04 & 0.15 & 0.31 & 0.06 & 0.26 & 0.24\\
text-babbage-001& 0.35 & 0.04 & 0.04 & 0.34 & 0.32 & 0.07 & 0.28 & 0.37\\
curie           & 0.50 & 0.01 & 0.04 & 0.23 & 0.49 & 0.01 & 0.22 & 0.30\\
text-curie-001  & 0.50 & 0.00 & 0.09 & 0.40 & 0.49 & 0.00 & 0.28 & 0.39\\
davinci         & 0.50 & 0.07 & 0.08 & 0.25 & 0.50 & 0.12 & 0.27 & 0.32\\
text-davinci-001& 0.51 & 0.07 & 0.08 & 0.38 & 0.51 & 0.14 & 0.19 & 0.39\\
text-davinci-002& 0.51 & 0.17 & 0.13 & 0.39 & 0.53 & 0.19 & 0.57 & 0.40\\
text-davinci-003& 0.53 & 0.52 & 0.33 & 0.54 & 0.57 & 0.30 & 0.80 & 0.43\\
GPT-3.5-Turbo   & 0.51 & 0.38 & \textbf{0.40} & 0.44 & 0.58 & 0.30 & 0.73 & \textbf{0.51}\\
GPT-4           & \textbf{0.55} & \textbf{0.60} & 0.31 & \textbf{0.74} & \textbf{0.67} & \textbf{0.42} & \textbf{0.91} & 0.46\\
\midrule
\midrule
GPT-4 (0-shot ICL)    & 0.60 & 0.19 & 0.25 & 0.72 & 0.65 & 0.27 & 0.85 & 0.48\\
GPT-4 (1-shot ICL)    & 0.66 & 0.24 & 0.30 & 0.70 & 0.71 & 0.38 & 0.78 & 0.41\\
GPT-4 (3-shot ICL)    & 0.61 & 0.70 & 0.70 & \textbf{0.75} & 0.69 & 0.29 & 0.70 & 0.40\\
GPT-4 (0-shot CoT)    & 0.57 & 0.57 & 0.28 & 0.73 & 0.66 & 0.43 & \textbf{0.90} & 0.53\\ 
GPT-4 (Manual CoT)    & \textbf{0.66} & \textbf{0.93} & \textbf{0.91} & 0.69 & \textbf{0.77} & \textbf{0.80} & 0.89 & 0.48\\  
GPT-4 (EF)            & 0.60 & -    & -    & 0.72 & 0.70 & -    & 0.87 & \textbf{0.53}\\
\bottomrule
\end{tabular} 
\caption{Performance (accuracy) of LLMs in causal tasks across the ladder of causation (Rung 1$\sim$3) on datasets including CLADDER, CaLM, CRASS, and E-CARE. The \colorbox{lightgray}{gray} column shows results for Rung 1 (corr=correlation), the white columns for Rung 2 (ATE=average treatment effect; CDE = controlled direct effect; BAJ= backdoor adjustment), and the \colorbox{LightCyan}{cyan} columns for Rung 3 (CR=counterfactual reasoning; NDE=natural direct effect; CE=causal explanation). The upper part shows results with basic prompts, while the lower part presents GPT-4 results with different prompting strategies. Data is sourced from \citet{chen2024causal} and \citet{jin2023cladder}.
}
\label{table:causal_effect_performance}
\vspace{-3mm}
\end{table*}

Unlike traditional causal discovery methods that rely solely on data values \cite{spirtes2000causation,spirtes2013causal,chickering2002learning}, LLMs can also leverage metadata (e.g., variable names, problem context) to uncover implicit causal relationships, making their reasoning closer to human cognition. Many studies have explored LLMs for causal discovery \cite{kiciman2023causal,cai2023knowledge,gao2023chatgpt,jin2023can,long2023can}, focusing on pairwise causal discovery and full causal graph discovery, often framed as multi-choice or free-text question-answering tasks.

\noindent\textbf{Pairwise causal discovery (PCD).} PCD focuses on inferring the causal direction ($A \rightarrow B$ or $A \leftarrow B$) or determining the existence of a causal relationship. \citet{kiciman2023causal} use variable names in prompts, showing that LLMs (e.g., GPT-3.5, GPT-4) outperform state-of-the-art methods on datasets like CauseEffectPairs \cite{mooij2016distinguishing} and domain-specific datasets like neuropathic pain \cite{tu2019neuropathic}. With proper techniques such as fine-tuning (as the example of fined-tuned Mistral shown in Table \ref{table: finetuning}), the performance of PCD can be improved significantly.
However, some studies \cite{zevcevic2023causal} suggest LLMs often act as "causal parrots", merely repeating embedded causal knowledge. Jin et al. \citeyearpar{jin2023can} proposes a correlation-to-causation inference (Corr2Cause) task, where LLMs performed close to random.  Although fine-tuning improved their performance, they still struggle with generalization in out-of-distribution scenarios. In summary, many studies \citet{gao2023chatgpt,kiciman2023causal,jin2023can,jin2023cladder,chen2024causal} take a nuanced stance, acknowledging LLMs' strengths in PCD tasks while highlighting their limitations in reliably determining the existence of causal relationships. 

\noindent\textbf{Full causal graph discovery.} Compared with PCD, identifying the full causal graph is a more complicated problem. In a preliminary exploration \cite{long2023can}, GPT-3 shows good performance in discovering the causal graph with 3-4 nodes for well-known causal relationships in the medical domain. In more complicated scenarios, the ability of different versions of GPT to discover causal edges \cite{kiciman2023causal} has been validated on the neuropathic pain dataset \cite{tu2019neuropathic} with 100 pairs of true/false causal relations. LLM-based discovery (GPT-3.5 and GPT-4) on Arctic sea ice dataset \cite{huang2021benchmarking} has comparable or even better performance than representative baselines including NOTEARS \cite{zheng2018dags} and DAG-GNN \cite{yu2019dag}. In \citet{ban2023query}, the combination of the causal knowledge generated by LLMs and data-driven methods brings improvement in causal discovery in data from eight different domains with small causal graphs (5$\sim$48 variables and 4$\sim$84 causal relations). However, similar to PCD, LLMs also face many debates about their true ability to discover full graphs \cite{zhou2024causalbench,jin2023can}.

%





\subsection{LLM for Causal Effect Estimation}
\label{sec:causal_effect}

Although comparatively underexplored, LLMs have also shown impressive performance in causal effect estimation. These works can be mainly categorized into two branches:
(1) \textbf{Causal effect in data:} LLMs estimate causal effects within data \cite{lin2023text,kiciman2023causal} by leveraging their reasoning capabilities and large-scale training data. CLADDER \cite{jin2023cladder} benchmarks LLMs for causal effect estimation tasks (e.g., ATE in Rung 2, and ATT, NDE, NIE in Rung 3). Although this task remains challenging, techniques like CoT prompting \cite{jin2023cladder} significantly improve performance.
(2) \textbf{Causal effect in models:} This branch investigates causal effects involving LLMs themselves, such as the impact of input data, neurons, or learning strategies on predictions \cite{vig2020investigating,meng2022locating,stolfo2022causal}. These studies help understand LLM behavior and support bias elimination \cite{vig2020investigating}, model editing \cite{meng2022locating}, and robustness analysis \cite{stolfo2022causal}. For example, \citet{stolfo2022causal} explores the causal effect of input (e.g., problem description and math operators) on output solutions in LLM-based mathematical reasoning. In \citet{vig2020investigating}, a causal mediation analysis for gender bias is conducted in language models.


\subsection{LLM for Other Causal Tasks}
\label{sec:other_causal}
Experiments \cite{chen2024causal,jin2023cladder,kiciman2023causal} have shown that there are various other causal tasks that LLMs can bring benefits to. 
\textbf{(1) Causal attribution}: LLMs show their capability in attribution tasks \cite{kiciman2023causal,cai2023knowledge} typically in the forms of "why" or "what is the cause" questions. Related tasks also include identifying necessary or sufficient causes \cite{liu2023trustworthy,kiciman2023causal}. By embedding human knowledge and cultural common sense, the results show that LLMs have the potential to flexibly address attribution problems in specific domains (such as law and medicine) where conventional methods may fall short \cite{kiciman2023causal}. 
\textbf{(2) Counterfactual reasoning}: Recent studies \cite{kiciman2023causal,jin2023cladder} explore different counterfactual reasoning scenarios, which are often in "what if" questions. While this task is one of the most challenging causal tasks, the improvement in LLMs compared to other methods is noteworthy.
\textbf{(3) Causal explanation}: Many recent works explore causal explanations with LLMs \cite{bhattacharjee2023llms,gat2023faithful,cai2023knowledge,gao2023chatgpt}. Despite ongoing debates regarding LLM's actual ability for causal reasoning, most empirical studies positively indicate that LLMs serve as effective causal explainers \cite{gao2023chatgpt}. Such achievement is powered by LLMs' capability of analyzing language logic and answering questions with natural language.

\subsection{Main Observations and Insights}
\vspace{-1mm}

From the evaluation above and results shown in Table \ref{table:causal_discovery_performance} $\sim$ Table \ref{table:causal_effect_performance}, we summarzie the main observations as follows: 
(1) \textbf{Model performance:} In general, many LLMs exhibit impressive performance in various causal tasks, especially in causal discovery, even with basic prompts. In some cases, their performance can be comparable to or even surpass human-level reasoning \cite{kiciman2023causal}. However, as the task difficulty increases from Rung 1$\sim$3, their performance becomes less satisfactory in higher-level complicated causal reasoning tasks \cite{chen2024causal}.
(2) \textbf{Enhancement through proper techniques:} The performance of LLMs can be significantly enhanced with effective prompting strategies (such as few-shot ICL and CoT) and other techniques like fune-tuning. These approaches enable models to be more causality-focused, with improved ability of leveraging context and adaptively following correct steps in different causal reasoning tasks. Additionally, these models can provide valuable insights through causal explanations.
(3) \textbf{General patterns:} While no definitive laws determine model performance universally, certain trends are still observable. For instance, scaling laws suggest that larger models generally perform better, although this is not always that straightforward. These trends provide valuable insights that can guide the future design and development of models.
(4) \textbf{Variability in model effectiveness:} There is currently no universally superior LLM or strategy for causal tasks, as their effectiveness can vary significantly depending on the specific scenario. These observations highlight the need for more nuanced and adaptable approaches. 
(5) \textbf{Common issues:} Current LLMs still struggle with many issues in causal tasks. For example, the answers often lack robustness and are sensitive to changes in prompts \cite{kiciman2023causal,jin2023cladder}. Furthermore, these models frequently default to memorizing and repeating information rather than actual causal reasoning \cite{zevcevic2023causal}, which can limit their effectiveness in complex causal scenarios. Besides, LLMs often fail to generate self-consistent causal answers, i.e., the answers from LLMs often conflict with each other. Ongoing debates about whether LLM truly performs causal inference also compel more in-depth analysis. 

\vspace{-1mm}
\section{Discussion and Future Prospects} 
\vspace{-1mm}
\label{sec:discussion}
In general, LLMs offer intriguing perspectives on causal inference, but current research also reveals many limitations, pointing to potential directions for future work that could advance the field \cite{zhang2023causality,kiciman2023causal}. 

\noindent\textbf{Involving human knowledge:} A more comprehensive  integration of human knowledge into LLMs can improve causal reasoning, enabling analysis across both general and specialized fields like finance, health, and law \cite{chen2024survey}.

\noindent\textbf{Improving data generation:} Real-world data often lacks verified causal relations and counterfactuals. LLMs can generate diverse, realistic data with reliable causal relationships, enriching datasets and improving causal reasoning model training.

\noindent\textbf{Addressing hallucinations:} In causal reasoning, hallucinations are commonly generated and difficult to detect, leading to misleading causal conclusions. Reducing them is essential to improve the reliability of LLM in causal tasks. 

\noindent\textbf{Improving explanation and interactivity:} Developing interpretable and instructable LLMs for causal reasoning is crucial. Techniques like fine-tuning, probing, prompt engineering, and optimizing reasoning chains can foster more collaborative and controllable causal inference.

\noindent\textbf{Exploring multimodal causality:} Real-world causal scenarios often involve multiple modalities. Recent studies have begun exploring causality across different modalities, such as images \cite{li2024multimodal} and videos \cite{lam2024causalchaos}. Future research could further investigate these multimodal approaches to enhance causal reasoning.

\noindent\textbf{Developing a unified causal benchmark:} There is currently a lack of unified and widely recognized benchmarks for evaluating causal performance for LLMs. Creating a comprehensive benchmark would facilitate LLM assessment.


\noindent\textbf{Advancing causality-specialized models:} Most current methods use original LLMs without sufficient focus on causality-centric model designs. There is a significant opportunity for further research and development in specialized causal LLMs to deepen their causal understanding and improve their effectiveness in causal inference.

\section{Limitations}
\vspace{-2mm}
In this survey, it is important to acknowledge certain limitations that shape the scope and focus of our review. First, our analysis is primarily centered on the application of LLMs for causal inference tasks, thereby excluding exploration into how causality is utilized within LLMs themselves. This decision provides a targeted perspective on leveraging LLMs to enhance causal inference but does not delve into the internal mechanisms or implementations of causal reasoning within these models.

Second, while we comprehensively examine the technical aspects and methodological advancements in using LLMs for causal inference, we do not extensively discuss ethical considerations or potential societal impacts associated with these applications. Ethical dimensions, such as fairness, bias mitigation, and privacy concerns, are critical in the deployment of AI technologies, including LLMs, and warrant dedicated attention and scrutiny in future research and applications. Addressing these limitations ensures a nuanced understanding of the opportunities and challenges in harnessing LLMs for causal inference while also advocating for responsible and ethical AI development and deployment practices.

 
\bibliography{ref}
\appendix

\end{document}